\documentclass[letterpaper, 10 pt, conference]{ieeeconf}  

\IEEEoverridecommandlockouts                              

\usepackage{graphics} 
\usepackage{xcolor}
\usepackage{epsfig} 
\usepackage{mathptmx} 
\usepackage{times} 
\usepackage{amsmath} 
\usepackage{amssymb}  
\usepackage{tipa}
\usepackage{algorithm} 
\usepackage{algpseudocode}   
\usepackage{lettrine}
\usepackage{graphicx}
\usepackage{etoolbox}
\usepackage{multirow} 
\usepackage{capt-of} 
\usepackage{cuted}
\usepackage{amsmath}
\usepackage{threeparttable}
\usepackage{setspace}
\let\oldtwocolumn\twocolumn
\renewcommand\twocolumn[1][]{%
    \oldtwocolumn[{#1}{
    \begin{center}
           \includegraphics[width=\linewidth]{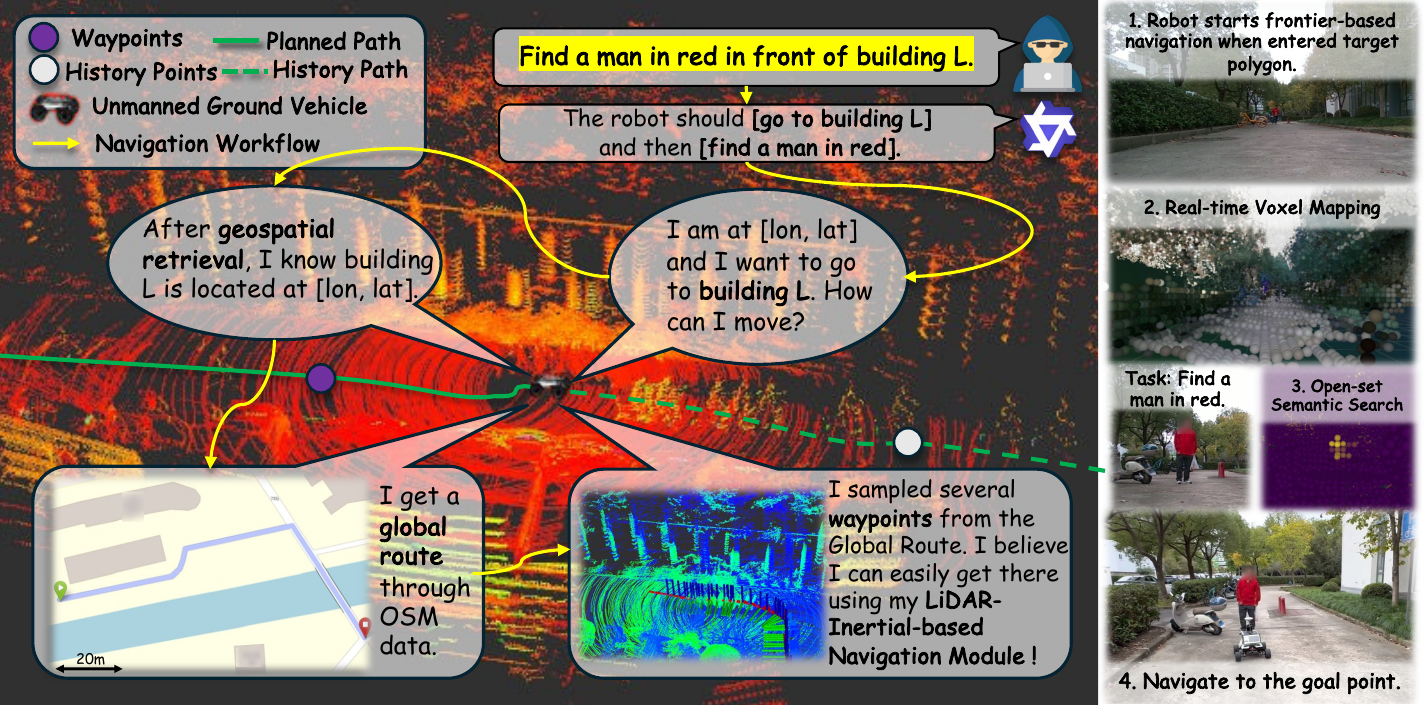}
           \captionof{figure}{We propose \textbf{G‑DRAGON}, a framework that accepts natural language commands and employs a Large Language Model (LLM) to parse instructions into navigation goals and object search targets. Global positioning is resolved by a generative retrieval module that queries an OpenStreetMap (OSM) database to generate an initial route. These global waypoints, combined with multi-modal sensor streams, guide a local planner for real-time execution. A closed-loop mechanism continuously feeds execution status back to the central planner for dynamic replanning. Upon arrival, the system transitions to an exploration mode to locate the specific target object.}
           \label{first}
        \end{center}
    }]
}

\begin{document}

\title{\LARGE \bf
G-DRAGON: Geospatial Reasoning and Dynamic Planning for Retrieval-Augmented Outdoor Navigation
}

\author{
Dongzhihan Wang$^{1,2}$,
Yi Du$^{2}$,
Jianan Sun$^{3}$,
Yuan Xue$^{1}$,
Yingchen Zhang$^{4}$, \\
Bing Xiao$^{5}$,
Chen Wang$^{2}$,
and Liang Xu$^{1,6,\dagger}$%
%
\thanks{$^\dagger$Corresponding author. The source code and datasets will be available at \textcolor{blue}{https://github.com/Anastasiawd/G-DRAGON.}}%
\thanks{The work is supported by the National Natural Science Foundation of China under Grant 62373239, 62333011, 62461160313, 62336005, and the State Key Laboratory of Space Intelligent Control under grant No. HTKJ2025KL502025. }%
\thanks{Part of this work was conducted while Dongzhihan Wang was an intern at the Spatial AI \& Robotics Lab, University at Buffalo.}%
\thanks{$^{1}$Dongzhihan Wang and Yuan Xue are with the School of Future Technology,
Shanghai University, Shanghai 200444, China.
}%
\thanks{$^{2}$Dongzhihan Wang, Yi Du, and Chen Wang are with the Spatial AI \& Robotics Lab,
Institute for Artificial Intelligence and Data Science,
Department of Computer Science and Engineering,
University at Buffalo, Buffalo, NY 14260, USA. Contact: \textcolor{blue}{https://sairlab.org.}}%
\thanks{$^{3}$Jianan Sun is with the College of Information Science and Technology,
Donghua University, Shanghai 200051, China.}%
\thanks{$^{4}$Yingchen Zhang is with the State Key Laboratory of AI Safety,
Institute of Computing Technology, Chinese Academy of Sciences,
and the University of Chinese Academy of Sciences, Beijing 100190, China.}%
\thanks{$^{5}$Bing Xiao is with the School of Automation,
Northwestern Polytechnical University, Xi'an 710072, China.}%
\thanks{$^{6}$Liang Xu is with the Hangzhou International Innovation Institute, Beihang University, Hangzhou 311115, China and is also with the School of Future Technology, Shanghai University, Shanghai 200444, China. Email: \textcolor{blue}{lxu006@e.ntu.edu.sg.}}%
}

\maketitle
\thispagestyle{empty}
\pagestyle{empty}



\begin{abstract}
Autonomous ground robots operating in large-scale outdoor environments require both robust long-range navigation and fine-grained ``last-mile'' exploration. 
Current advances in visual-language navigation (VLN) work well at short-range tasks, lacking geospatial grounding for long-distance missions. 
Some OpenStreetMap (OSM)-based methods relying on cloud-based Large Language Models (LLMs) are prone to factual hallucination and cannot conduct ``last-mile'' exploration based on human instruction. 
To address these challenges, we present G-DRAGON, a retrieval-augmented framework for outdoor, open-world navigation.
This framework maps natural-language commands to versioned, local OSM entities via generative retrieval based on  lightweight LLM, yielding accurate coordinates for global route planning. 
A high-level planning module bridges global topological routes with the SLAM system, projecting geospatial waypoints into the robot's navigable frame. 
For the ``last mile," the framework transitions to frontier-based exploration and open-set semantic voxel mapping to localize open-vocabulary targets. 
Experimental results in simulation demonstrate our framework outperforms state-of-the-art baselines.
Furthermore, we validate the system in unseen real-world urban environments on an Unmanned Ground Vehicle (UGV), successfully completing person-search missions with trajectories of up to 500m.
\end{abstract}

\section{INTRODUCTION}
\lettrine{A}{utonomous} ground robots are increasingly tasked with complex missions in large-scale outdoor environments, ranging from campus delivery to safety patrols. 
A typical instruction, such as \textit{``Navigate to the library and find the person in a red jacket"}, requires the robot to first traverse long distances across complex campus environments and then shift to close-range semantic search in the “last mile.” Long-distance navigation in urban terrains prioritizes efficiency by operating on coarse environmental priors, while last-mile exploration instead requires fine-grained perception for identifying open-vocabulary targets.

Existing methods often struggle to address these dual challenges. End-to-end navigation methods, ranging from goal-conditioned policies \cite{sridhar2024nomad, shah2023vint} to vision-language-action (VLA) models applied in visual-language navigation (VLN) \cite{navfom, cheng2024navila, liu2025trackvla++}, excel in short-horizon tasks but often struggle with long-distance navigation. Since they rely solely on visual inputs without geospatial grounding, they are prone to getting lost in unstructured outdoor environments and cannot generate reliable plans for long-distance missions. While OpenBench \cite{open} provides a solution in such scenario, it suffers from grounding latency and hallucination inherent to cloud-based LLMs, and lacks the flexibility to handle targets beyond static OSM addresses or doors.

To bridge these gaps, we propose \textbf{G-DRAGON} (\textbf{G}eospatial Reasoning and \textbf{D}ynamic Planning for \textbf{R}etrieval-\textbf{A}u\textbf{G}mented \textbf{O}utdoor \textbf{N}avigation), a unified framework designed for large-scale navigation and ``last-mile" exploration that decouples high-level reasoning from low-level control. 
Crucially, instead of relying on cloud-based LLMs, we leverage locally deployed models to handle reasoning tasks. We formulate the localization of natural language targets as a \textit{generative retrieval} problem \cite{tay2022transformermemorydifferentiablesearch}, \cite{decao2021autoregressiveentityretrieval}, mapping semantic aliases i.e., ``library" to concrete OSM entities with geodetic coordinates. 
This output informs a high-level planner to parse tasks via Behavior Trees and synthesize global paths for downstream navigation modules. Subsequently, the system employs environmental feedback to generate local trajectories that track the global reference, activating an open-vocabulary search mode upon reaching the target's vicinity.

We conduct comprehensive evaluations of the entire system in campus-level simulation environments and the results show that we outperform baselines by 50\%. Furthermore, we demonstrate the real-world feasibility of our approach by completing 4 long-range navigation trials on a mobile robot. The main contributions of this work are summarized as follows:

\begin{itemize}
    \item \textbf{GeoQA for Global Route Generation:} We design a geospatial Question-Answering (QA) module based on generative retrieval to align natural language commands with appropriate OSM entities. It precisely outputs latitude and longitude coordinates, which are passed to the downstream global planner to initialize the route.
    
    \item \textbf{RAPPER} (\textbf{R}etrieval-\textbf{A}ugmented \textbf{P}lanner with \textbf{P}rogrammatic and \textbf{E}xecutable \textbf{R}easoning): We propose a hierarchical planner which parses tasks from natural language, queries \textbf{GeoQA} and synthesizes Behavior Trees through a locally deployed LLM. By constraining the LLM to structured programmatic outputs, the system safely generates executable task planning and global path planning.
    
    \item \textbf{N\textipa{\ae}VIS (Navigation and Exploration System):} We present a seamless integration of long-range navigation and ``last-mile" exploration. 
    It employs frontier-based navigation coupled with open-set semantic voxel mapping to robustly localize the specific target objects mentioned in the user's command.
\end{itemize}

\begin{figure*}[t!]
\vspace{5pt}
    \centering
    \includegraphics[width=1.0\linewidth]{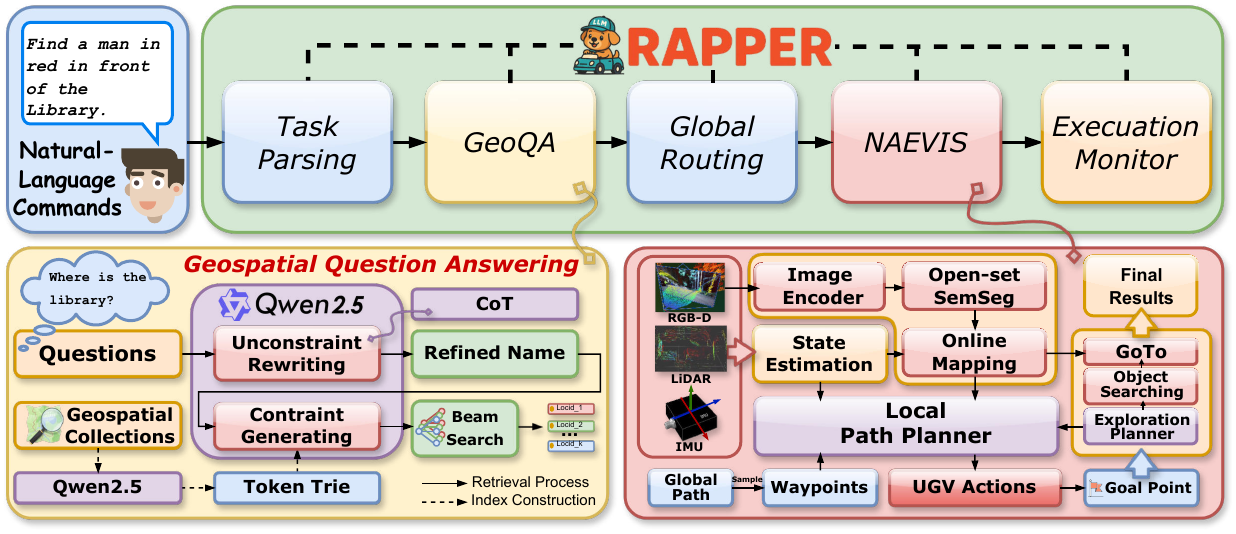}
    \caption{The architecture of \textbf{G-DRAGON}. The system processes natural language instructions within a local deployed environment. The \textbf{GeoQA} module retrieves OSM metadata to ground semantic targets from abstract user commands into precise geodetic coordinates. The \textbf{RAPPER} module serves as the reasoning orchestrator, leveraging a locally deployed Qwen2.5 LLM to reason and generate task-and-motion planning. The \textbf{N\textipa{\ae}VIS } module executes the generated task planning by fusing multi-modal sensor data, handling long-range navigation and open-vocabulary object search in the ``last-mile".}
    \label{fig:pipeline}
\vspace{-10pt}
\end{figure*}


\section{Related Work}

\subsection{Long-range Language-Guided Outdoor Navigation}

Driven by foundation models, language-guided navigation has advanced through goal-conditioned methods \cite{sridhar2024nomad}, \cite{shah2023vint} and VLA \cite{navfom} \cite{cheng2024navila}, \cite{liu2025trackvla++} frameworks. However, these vision-centric approaches remain largely confined to short-horizon or indoor settings. To address long-range outdoor navigation, recent works \cite{lmnav} and \cite{viking} integrate topological graphs or aerial hints with pre-trained models. Yet, they rely heavily on visual correspondence, limiting their reliability in repetitive environments or under varying lighting conditions in large-scale open worlds, unlike approaches grounded in explicit geospatial data.

\subsection{Geospatial Reasoning and OSM-Based Navigation}
To bridge semantic instructions with actionable targets, recent works utilize a hierarchical architecture \cite{sayplan} to decouple task planning from control. However, existing methods lack robustness for autonomous ground systems. GeoNav \cite{xu2025geonav} is designed for aerial perspectives, rendering it unsuitable for ground-level path planning. Embodied-RAG \cite{embodied-rag} supports retrieval but confines its search to historical traversal data rather than global semantic locations. Additionally, frameworks like OpenBench \cite{open} rely entirely on cloud-based LLMs i.e. GPT-4o, which prevents fully offline operation and suffers from hallucinations in large-scale spatial reasoning tasks.

\subsection{``Last-Mile” Exploration}

In “last-mile” scenarios, instead of exploring continuously, the system activates exploration only when the robot approaches the destination. It requires open-vocabulary perception for natural language queries, robustness in outdoor environments, and crucially, the ability to run on an onboard edge device. Current methods are limited in bridging these gaps. Traditional frontier methods \cite{yamauchi} maximize coverage but remain semantically blind. Conversely, recent methods like SG-Nav \cite{sgnav} and ApexNav \cite{apexnav} rely heavily on cloud-based inference and disigned for indoor scenarios. While OpenBench \cite{open} explicitly proposes a solution for such last-mile navigation, its lightweight visual encoders and flat 2D projection limit its open-vocabulary exploration performance in complex outdoor environments.
\section{Methods}

To support robust query-driven outdoor navigation on resource-constrained ground robots, we propose G-DRAGON, an end-to-end framework consisting of three modules, as shown in Fig.~\ref{fig:pipeline}. GeoQA grounds natural-language queries in OSM and returns geodetic targets. RAPPER parses the task and generates a global route. N\textipa{\ae}VIS executes long-distance outdoor navigation and, near the target region, performs open-vocabulary search to localize the final object. The following sections describe these modules and the real-time implementation details.

\subsection{GeoQA Module} 

\subsubsection{Task Statements}

We formulate natural-language grounding as a retrieval task: given a query $q$, the goal is to select an entity $e^{*}$ from the entity set $\mathcal{E}$ of the OSM-derived knowledge base $\mathcal{K}$ that best matches both semantic intent and geospatial constraints, yielding precise coordinates for downstream planning. This is non-trivial due to three factors: (1) queries are open-ended and semantically diverse, making sparse lexical matching insufficient; (2) data is limited and scene-specific, limiting the viability of dense retrievers that require large-scale training; and (3) the task demands selecting one concrete entity, while free generation often drifts and cannot guarantee valid outputs.

To address this issue, we adopt a \textbf{``zero-shot generative retrieval''} framework. This approach fully leverages the LLM’s understanding and generation capabilities, while imposing constraints on its decoding process so that the output is guaranteed to fall within the candidate set. In this way, it simultaneously resolves both the dependence on training and the problem of output drift, making it highly suitable for our scenario.

\subsubsection{GeoQA}
As shown in Fig.~\ref{fig:pipeline}, our method depends on only a single pretrained LLM. We first assign each entry a unique and highly discriminative identifier (ID), and then construct decoding constraints from the set of all valid IDs. Finally, we invoke the same LLM twice (once for free-form reasoning and once for constrained decoding) to complete the retrieval process. 

We first create a unique and highly discriminative natural-language ID for each entity. Specifically, for each entry, we (1) use the entry name as a low-level identifier; (2) assign it a coarse-grained category name as a high-level identifier; and (3) concatenate the high-level ID and the low-level ID with a hyphen to obtain the final ID (e.g., ``\textit{Teaching Area-Building A}''). This hierarchical naming scheme not only reduces ambiguity among entities that share similar base names, but also aligns with natural human expressions, making it easier for LLMs to reason about and reliably generate the identifiers.

Next, we build a prefix tree (Trie) over the entity IDs similar to~\cite{decao2021autoregressiveentityretrieval} to guarantee valid retrieval. During inference, this Trie acts as an indexing constraint on the decoder: at each generation step, the model’s next-token distribution is masked so that only tokens corresponding to valid child nodes in the Trie remain admissible. Formally, for decoding step $t$ we define the Trie-constrained distribution as
\begin{equation} \label{eq:trie_constrained_prob}
P_{\mathcal{T}}(y_t \mid y_{<t}, q) =
\begin{cases}
  P(y_t \mid y_{<t}, q), & \text{if } y_t \in \text{children}(y_{<t}) \\
  0, & \text{otherwise},
\end{cases},
\end{equation}
where $\text{children}(y_{<t})$ denotes the valid next tokens given the current prefix in the Trie $\mathcal{T}$. This mechanism effectively eliminates semantic hallucination by ensuring that every generated token sequence corresponds to an extant entity in the OSM database $\mathcal{K}$.

At inference time, GeoQA proceeds in two stages. (1) Given a query $q$, we first apply a lightweight Chain-of-Thought (CoT) prompt~\cite{wei2023chainofthoughtpromptingelicitsreasoning} to extract structured reasoning cues. The LLM is instructed to identify intermediate variables such as the target category $\hat{c}$ (e.g., \textit{Teaching Area}) and salient modifiers (e.g., \textit{Building A}). These intermediate variables serve as semantic priors that clarify the intended region of the map and narrow down the set of plausible candidates. (2) Conditioned on the extracted cues and the original query, we then perform constrained beam search~\cite{scholak-etal-2021-picard} to generate the target entity ID. At each decoding step, token expansion is restricted by the Trie as in Eq.~\eqref{eq:trie_constrained_prob}, so that the beam only explores paths consistent with valid entity identifiers. In effect, the model \emph{freely reasons} about relevance in the CoT stage, and then \emph{selects} a concrete entity through Trie-constrained generation. This two-stage process eliminates free-form hallucination and guarantees that the final output is a unique, verifiable entity in $\mathcal{K}$, enabling seamless integration with downstream navigation planners.



\begin{algorithm}[t]
\caption{The RAPPER Module}
\label{alg:rapper}
\setstretch{1.0} %
\begin{algorithmic}[1]
\Require Instruction $I$, pose $P_{curr}$, map $\mathcal{M}$, skill library $\mathcal{S}$
\Ensure Mission Status (\textbf{Success} / \textbf{Failure})

\State $\mathcal{T} \gets \textsc{LLM\_Parse}(I)$ 
\State $\text{BT} \gets \textsc{Build\_BT}(\mathcal{T}, \mathcal{S})$ 
\While{$\text{BT}$ is \textbf{Running}}
    \State $n \gets \textsc{Tick}(\text{BT})$
    \If{$n.\text{skill} = \textsc{GlobalPlanning}$}
        \State $P_{poly} \gets \textsc{GeoQA}(n.\text{target},\, \mathcal{M})$
        \State $\tau \gets \textsc{OSRM}(P_{curr},\, P_{poly},\, \mathcal{M})$
        \State $s \gets \textbf{Success}$
    \ElsIf{$n.\text{skill} = \textsc{Navigation}$}
        \State $s \gets \textsc{FollowPath}(\tau)$ 
    \ElsIf{$n.\text{skill} = \textsc{Exploration}$}
        \State $s \gets \textsc{GuideTo}(\text{N\ae VIS},\, n.\text{target})$
    \EndIf
    \State \Call{UpdateBT}{$\text{BT},\, s$}
\EndWhile
\State \Return $\text{BT}.\text{status}$
\end{algorithmic}
\end{algorithm}

\subsection{RAPPER}
RAPPER is a hierarchical control framework that serves as the central reasoning orchestrator of the G-DRAGON system, built upon a Qwen2.5-14B backbone LLM. As described in Algorithm \ref{alg:rapper}, it utilizes a locally deployed LLM to parse instructions into a Behavior Tree, decomposing tasks and selecting skills from the predefined pool. RAPPER then generates global routes from the specified targets and transmits executable decisions to downstream executors, while dynamically monitoring the robot’s state using real-time perception and planner feedback. Unlike conventional end-to-end agents whose iterative cloud-based queries introduce high latency and increase the risk of hallucination, RAPPER decouples high-level semantic reasoning from low-level execution entirely \textbf{within a local deployment}. Through fully offline inference, RAPPER dynamically builds a Behavior Tree (BT) \cite{bt} in a single shot. The BT generation process is constrained by a predefined action and condition space, ensuring every node corresponds to an executable operation.

\subsubsection{Task Grounding}
Upon receiving a natural language instruction $\mathcal{I}$, RAPPER employs the locally deployed LLM to parse the user's intent, decomposing $\mathcal{I}$ into a logical sequence of sub-tasks: $\mathcal{T} = \{T_{nav}^{(1)}, T_{exp}^{(1)}, \dots, T_{nav}^{(k)}\}$. Here, $T_{nav}$ and $T_{exp}$ correspond to navigation tasks \textit{``Where to go''} and last-mile semantic exploration tasks \textit{``What to find''}, respectively. For each $T_{nav}$, RAPPER queries the \textbf{GeoQA} module to retrieve geographic coordinates. To ensure precise navigation goal assignment, we compute the target goal $P_{goal}$ by projecting the robot's position onto the nearest point of the target's polygon contour, rather than using the geometric centroid. This refined coordinate is then fed into a local routing engine \cite{osrm} based on OSM nodes to generate the initial global topological path $\tau_{global}$.

\subsubsection{Behavior Tree Construction and Execution} 
To accommodate diverse and unstructured instructions, RAPPER utilizes the locally deployed LLM to dynamically construct a BT. Leveraging the exceptional code generation capabilities of modern LLMs, we formulate the BT construction as a structured code synthesis task, inspired by the ProgPrompt paradigm \cite{singh2022progpromptgeneratingsituatedrobot}. We provide the LLM with a prompt context that defines the robot's action space as a \textbf{skill pool} of primitives i.e., \texttt{GlobalPlanning}, \texttt{FollowPath}. Through a Chain-of-Thought (CoT) process, the model first explicitly reasons about the logical dependencies within $\mathcal{T}$, and then maps these sub-tasks to a hierarchical XML-formatted structure. We elect XML as the intermediate representation due to its strict syntactic structure, which naturally aligns with the hierarchical nature of BTs. This allows RAPPER to generate complex control logic, such as Sequences and Fallbacks, directly in an robot-executable format. 

For execution, the instantiated XML tree is parsed by a runtime engine that cyclically ``ticks'' the nodes, translating abstract tags into specific action messages. To ensure robustness during execution, the system continuously monitors the robot's real-time state. If the robot remains stationary due to unexpected blockages or planner failures, the BT's inherent structure triggers a recovery mechanism to re-initialize the path search, ensuring the logical plan is physically realized without high-level human instructions. The system enters a continuous monitoring loop, cyclically executing the BT nodes. This process iterates until the entire tree traverses to a completion state, at which point the final task status is returned to the user, marking the end of the mission.

\subsubsection{Transition to Last-Mile Exploration} Once the navigation task $T_{nav}$ returns \texttt{Success}, RAPPER seamlessly transitions the context to $T_{exp}$. It dispatches the semantic target description i.e., \textit{``a man in red''} to the downstream exploration module in \textbf{N\textipa{\ae}VIS}. RAPPER then enters a monitoring state, processing real-time feedback until the target is localized or the search area is exhausted.

\subsection{N\textipa{\ae}VIS}



    
This module processes multi-modal inputs from an RGB-D camera, a LiDAR, and an IMU, fusing them into a consistent spatial representation for navigation. We leverage a LiDAR-inertial odometry (LIO) \cite{zhao2021super}, \cite{xu2021fastlio2fastdirectlidarinertial} to generate real-time state estimates, while RGB-D data is concurrently processed by RayFronts \cite{alama2025rayfrontsopensetsemanticray} to update an open-set semantic voxel map.

\subsubsection{Heading Initialization and Waypoints Sampling}
To bridge the domain gap between the global geodetic coordinates and the robot's local odometry, we implement a dynamic initialization maneuver. Once the GPS signal stabilizes, the robot traverses a straight-line trajectory of length $d_{init} = 5m$. We derive the global heading vector $\mathbf{v}_{global}$ based on the filtered GPS coordinates of the start ($\mathbf{p}_{start}$) and end ($\mathbf{p}_{end}$) points:
\begin{equation}
\mathbf{v}_{global} = \mathbf{p}_{end} - \mathbf{p}_{start} = [\Delta x_{ENU}, \Delta y_{ENU}]^T.
\end{equation}
The deviation angle $\theta_{bias}$ between the robot's current heading (local $x$-axis) and the Magnetic East is computed as:
\begin{equation}
\theta_{bias} = \operatorname{atan2}(\Delta y_{ENU}, \Delta x_{ENU}).
\end{equation}
Consequently, any global waypoint $W_{global}^{(i)}$ from the routing engine \cite{osrm} is transformed into the robot's local navigation frame $W_{local}^{(i)}$ via the rotation matrix $\mathbf{R}(\theta_{bias})$:
\begin{equation}
\begin{split}
W_{local}^{(i)} &= \mathbf{R}(\theta_{bias})^T \left( W_{global}^{(i)} - \mathbf{p}_{end} \right), \\
\mathbf{R}(\theta) &= \begin{bmatrix} \cos\theta & -\sin\theta \\ \sin\theta & \cos\theta \end{bmatrix}.
\end{split}
\end{equation}
This ensures that the route remains fixed relative to the robot's initial orientation.

Following this, to balance local planning efficiency with tracking accuracy, we propose a non-uniform, curvature-adaptive sampling strategy for the global path $\tau_{global}$. Let the path be represented as a sequence of discrete nodes $\{n_1, n_2, \dots, n_k\}$. For each segment, we evaluate the local path curvature $\kappa_i$ to differentiate straight roads from intersections, approximated by the change in heading angle $\Delta \phi$ over unit distance. The sampling interval $\delta_s^{(i)}$ for the $i$-th segment is dynamically determined by:
\begin{equation}
\delta_s^{(i)} = 
\begin{cases} 
\delta_{fine} & \text{if } \kappa_i > \kappa_{thresh} \quad \\
\delta_{coarse} & \text{otherwise} \quad.
\end{cases}
\label{eq:adaptive_sampling}
\end{equation}
In our implementation, we set the fine sampling interval $\delta_{fine} = 3m$ to ensure smooth maneuverability at corners, and the coarse interval $\delta_{coarse} = 20m$ to reduce the computational load on the local planner \cite{yang2022farplannerfastattemptable}, \cite{aede} during straight-line navigation.

\subsubsection{``Last-Mile" Exploration}
Upon reaching the vicinity of the target, N\ae VIS transitions to a coarse-to-fine semantic exploration mode. First, the robot performs an in-place $360^{\circ}$ scan. This initialization step rapidly populates the local semantic voxel map, checking for immediate target visibility and generating an initial set of frontiers. If the target is not detected during the scan, the system proceeds to  a  frontier-based exploration mode. We adapt Yamauchi’s  strategy \cite{yamauchi} by imposing a hard spatial constraint: candidate frontiers are first extracted from the global occupancy grid and then strictly filtered via a point-in-polygon (PiP) test against the target's OSM boundary. This ensures the robot navigates greedily to the nearest valid frontier without drifting into irrelevant areas. 

Concurrently, similar to \cite{kim2025raven}, the image encoder performs real-time, query-conditional semantic segmentation on the voxel map. Once voxels exceeding a semantic similarity threshold are detected during either the initial scan or the subsequent exploration, the system interrupts the search. It then transitions to the object navigation phase, where high-relevance voxels are clustered using DBSCAN. Finally, the centroid of the dominant cluster is projected onto the navigable ground plane to serve as the final 2D navigation goal.

\section{Experiments}

\begin{table*}[!t]
\centering 
\caption{Quantitative Comparisons on Long-Range Navigation and ``Last-Mile" Exploration.}
\label{tab:sim_results}
\small 
\def\arraystretch{1} 
\setlength{\tabcolsep}{15pt} 
\begin{tabular}{ccccccccc}
\hline
\multirow{2}{*}{Method} & \multicolumn{2}{c}{Task 1 (300m)} & \multicolumn{2}{c}{Task 1 (800m)}  & \multicolumn{2}{c}{Task 2 (300m)} & \multicolumn{2}{c}{Task 2 (800m)}\\
\cline{2-3} \cline{4-5} \cline{6-7} \cline{8-9}
    & SR & SPL  & SR & SPL  & SR & SPL  & SR & SPL  \\
\hline
    NoMaD \cite{sridhar2024nomad} & 33.33 & 16.84 & 0 & 0 & 11.11 & 4.26 & 0 & 0  \\
    OPEN \cite{open} & 77.78 & 42.56 & 26.67 & 12.42 & 20 & 10.53 & 17.78 & 9.42  \\
\textbf{Ours} & \textbf{100} & \textbf{85.25} & \textbf{95.56} & \textbf{81.45} & \textbf{80} & \textbf{38.52} & \textbf{62.22}  & \textbf{35.67}\\
\hline
\vspace{-15pt}
\end{tabular}
\end{table*}


\subsection{Experiment Setup}
\subsubsection{Dataset Construction for Geospatial Reasoning}

To rigorously evaluate the performance of our GeoQA module, we constructed a dataset comprising 5,000 natural language queries. These queries were synthesized using an external LLM based on approximately 300 building nodes extracted from a $1 \times 1$ km$^2$ region centered on a large-scale university campus. To assess robustness, the dataset is stratified into two difficulty levels with a 7:3 ratio. The \textbf{Easy (Explicit)} subset consists of queries with direct destination descriptions i.e., \textit{“Where is the library?”}. In contrast, the \textbf{Hard (Implicit)} subset challenges the system with functional descriptions i.e., \textit{“Find a place to eat”} or queries containing distractors i.e., \textit{“Not Building A, but Building B”}.

\subsubsection{Simulation Environment}
\begin{figure}[t]
\vspace{5pt}
    \centering
    \includegraphics[width=0.95\linewidth]{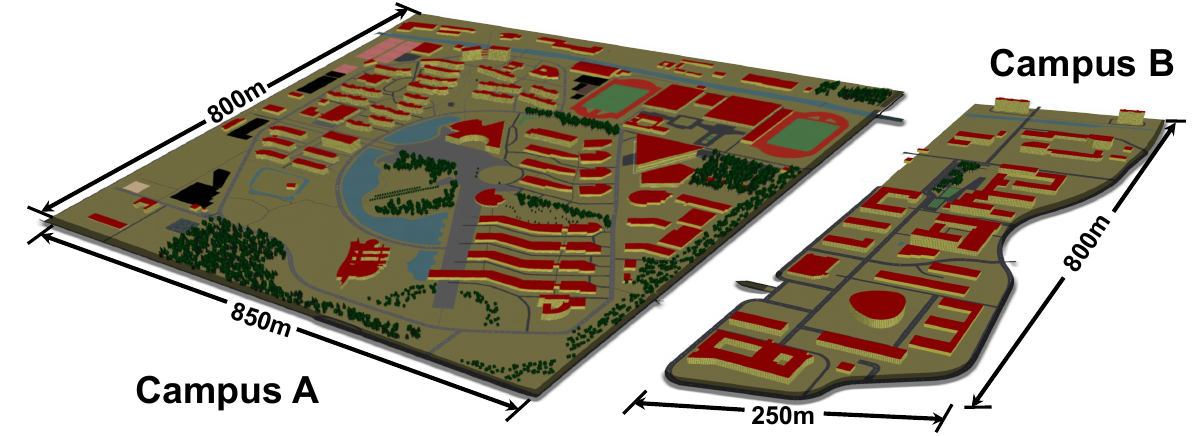}
    \caption{Simulation environment for N\textipa{\ae}VIS system.}
    \label{fig:sim_env}
\vspace{-10pt}
\end{figure}
To validate the full G-DRAGON navigation pipeline, we developed simulation scenarios grounded in real-world geospatial data. We constructed two large-scale campus environments within NVIDIA Isaac Sim, shown in Fig. \ref{fig:sim_env} through a three-stage pipeline: (1) extraction of raw topological data from OSM; (2) editing to experimental constraints; and (3) conversion of maps into 3D models for import into the simulator. An NVIDIA Nova Carter UGV was deployed as the robotic agent to execute navigation tasks within these digital twins. Experiments utilized an NVIDIA RTX 5090 GPU to handle simulation rendering. We synthesized GNSS measurements by transforming simulation ground-truth poses into the WGS-84 frame via a fixed ENU projection anchored to the real-world campus origin. To emulate sensor realism, Gaussian noise was superimposed on the signals before publishing them to the navigation stack.
\subsubsection{Real-Robot Setup}

We evaluate our system on an Agilex Scout Mini UGV shown in Fig. \ref{fig:realrobot}(a), equipped with an NVIDIA Jetson AGX Orin. The perception suite includes a 16-beam LiDAR, an RGB-D camera, an IMU, and a GNSS module.

\subsection{Evaluation Metrics}
To comprehensively evaluate both the geospatial reasoning and the navigation performance, we adopt the following metrics:

\subsubsection{Retrieval Metrics}
We evaluate the generative retrieval performance using \textbf{Recall@K} ($R@K$). It measures the proportion of test queries where the ground-truth OSM entity identifier appears in the top-$K$ generated candidates. For a dataset of $N$ queries, let $y_i$ be the ground-truth entity and $\hat{Y}_{i, K}$ be the set of top-$K$ predictions for the $i$-th query. $R@K$ is defined as:
\begin{equation}
    R@K = \frac{1}{N} \sum_{i=1}^{N} \mathbb{I}(y_i \in \hat{Y}_{i, K}),
\end{equation}
where $\mathbb{I}(\cdot)$ is the indicator function. 

\subsubsection{Navigation Metrics}
Following standard navigation benchmarks \cite{anderson2018evaluationembodiednavigationagents}, we evaluate performance using two metrics: \textbf{Success Rate (SR)} and \textbf{Success weighted by Path Length (SPL)}. SR reports the percentage of episodes where the agent successfully arrives within a predefined threshold of the target. SPL further qualifies this success by measuring navigation efficiency, scaling the success indicator by the ratio of the optimal path length to the actual traversed distance. Since our system operates end-to-end, a high SR inherently reflects the correctness of the high-level task planning, as incorrect planning would lead to navigation failure.
\subsection{Comparison Results of Geospatial Reasoning}

\begin{table}[t]
\caption{Comparison of Geospatial Reasoning Performance.}
\vspace{-10pt}
\label{tab:geoqa_results}
\begin{center}
\def\arraystretch{1.0}%
\setlength{\tabcolsep}{10pt}
\begin{tabular}{ccccc}
\hline
\multirow{2}{*}{Method} & \multicolumn{2}{c}{Easy Split} & \multicolumn{2}{c}{Hard Split}  \\
\cline{2-3} \cline{4-5}
    & R@1 & R@5 & R@1 & R@5  \\
\hline
    BM25 \cite{beam25} & 90.0 & 96.3 & 33.3 & 71.7  \\
    DPR \cite{dpr}& 44.6 & 55.2 & 16.2 & 25.3  \\
\hline
    Qwen2.5-14B & 87.8 & 92.4 & 67.7 & 72.7 \\
    GPT-4o & \underline{96.7} & \underline{97.5} & \textbf{87.4} & \textbf{88.1}  \\
\hline
\textbf{GeoQA} & \textbf{97.6} & \textbf{99.6} & \underline{74.8} & \underline{81.8} \\
\hline
\end{tabular}%
\vspace{-20pt}
\end{center}
\end{table}

To evaluate retrieval performance, we benchmark GeoQA against sparse retriever BM25 \cite{beam25}, dense retriever DPR \cite{dpr} and foundation models (Vanilla Qwen 2.5 14B, GPT-4o). Table \ref{tab:geoqa_results} indicates that traditional retrievers yield suboptimal performance as generic embeddings fail to capture campus-specific topology, while unconstrained LLMs suffer from severe hallucinations and format instability.

Notably, in the ``\textit{Easy}" split, GeoQA surpasses even GPT-4o by leveraging constrained decoding to eliminate formatting errors. While GPT-4o exhibits superior reasoning in the ``\textit{Hard}" split, its reliance on cloud connectivity renders it unsuitable for field robotics. In contrast, GeoQA emerges as the only viable solution for strictly offline deployment. By significantly outperforming all local baselines, it establishes itself as the indispensable choice for connectivity-denied environments, offering the optimal trade-off between autonomy and reasoning capability.



\subsection{Quantitative Results on Simulation Environments}

\begin{figure*}[t]
\vspace{5pt}
      \centering
      \includegraphics[width=1.0\linewidth]{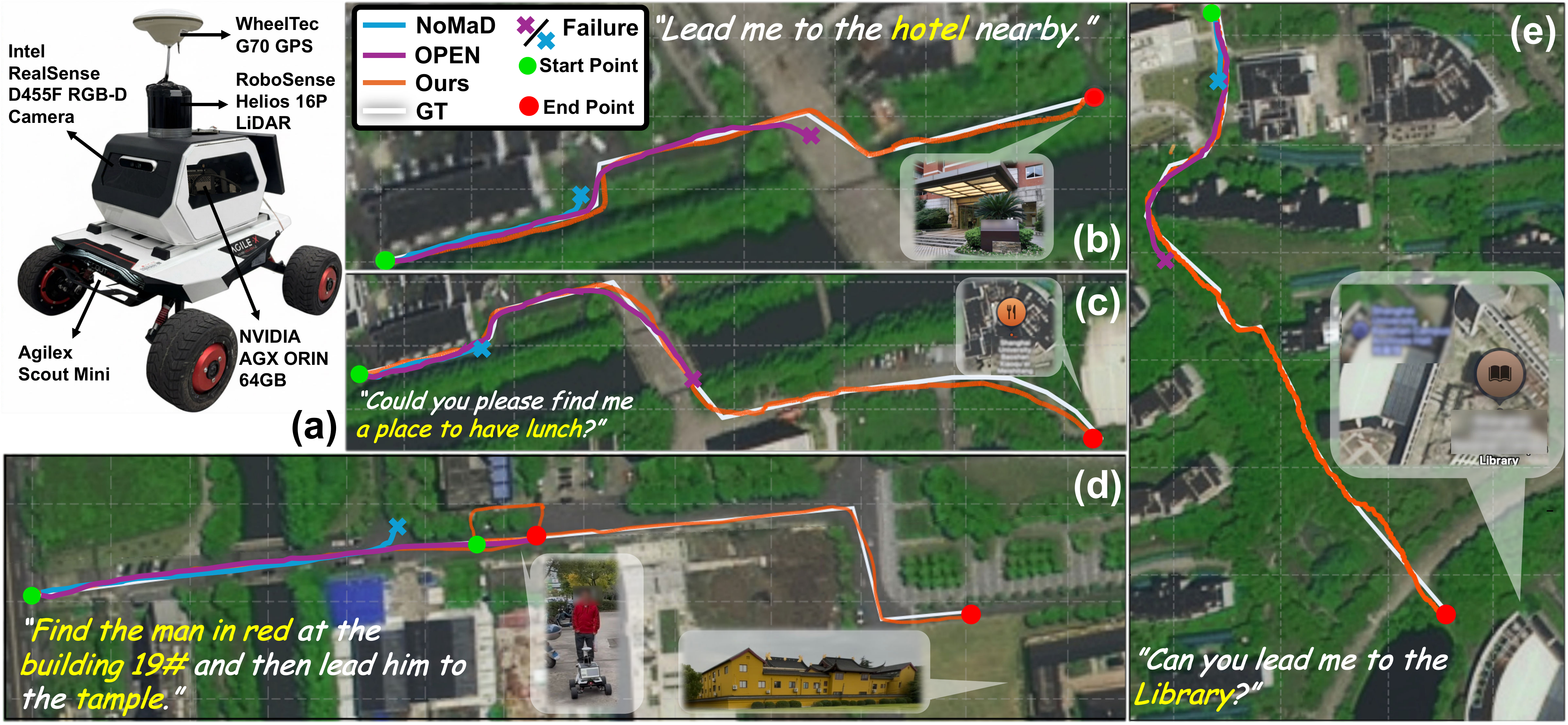}
      \caption{Illustration of our real-robot setup and real-world experimental results. (a) We deploy our system on a UGV. (b)–(e) present the tasks designed to evaluate the N\textipa{\ae}VIS system and the corresponding results.}
      \label{fig:realrobot}
\end{figure*}

\begin{figure*}[t!]
      \centering
      \includegraphics[width=1.0\linewidth]{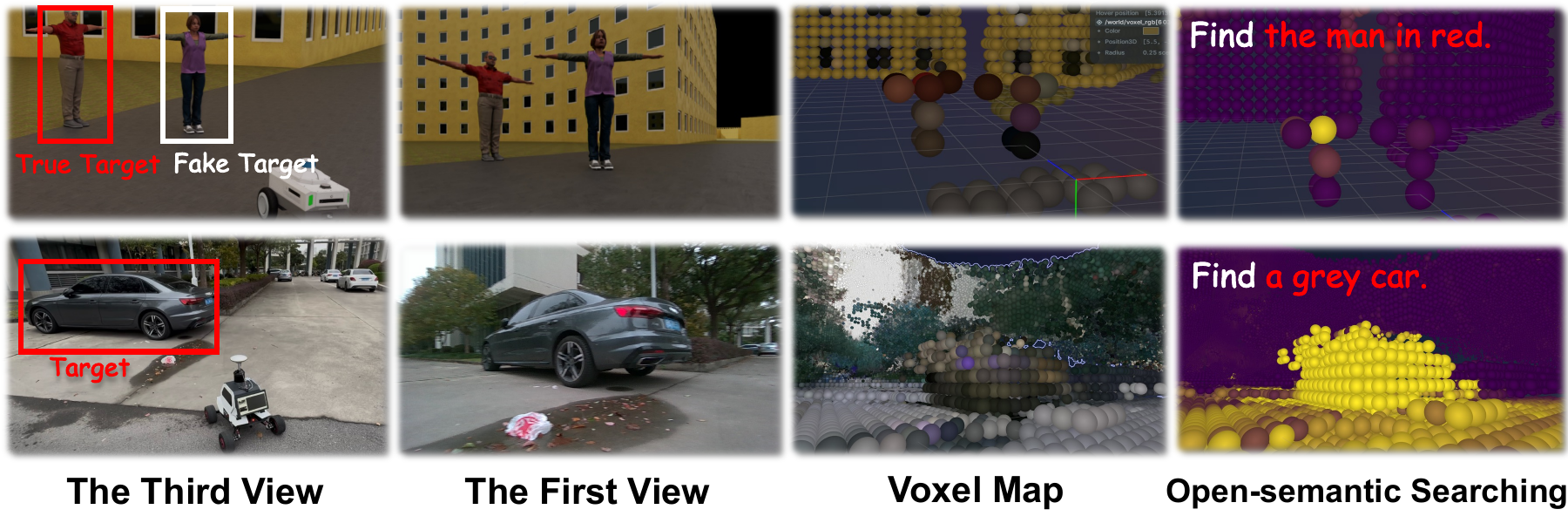}
      \caption{Visualization of ``Last-Mile" Exploration in Simulation (Top) and Real-world (Bottom). Columns show the 1st-person view, 3rd-person view, voxel map and open-semantic segmentation. The heatmap highlights the target object. Notably, our method successfully isolates the correct target while ignoring distractors.}
      \label{fig:rayfronts}
\vspace{-15pt}
\end{figure*}

We selected NoMaD and OPEN as baselines because they satisfy our constraints of open-source availability, offline edge deployment.
For fairly evaluation, we provide OPEN with exact coordinates and NoMaD with target images, as RAPPER naturally avoids planning failures via OSM grounding. Two tasks are evaluated at 300m and 800m. Each range includes 45 diverse episodes i.e., $5 \text{ buildings} \times 3 \text{ positions} \times 3 \text{ yaws}$ to comprehensively test initial heading and spatial variations. Task 1 focuses on building-goal navigation, where reaching within 5 m of the target constitutes success. Task 2 extends this by requiring subsequent exploration around the building polygon to locate a specific object, with a stricter success threshold of 2 m.

Table \ref{tab:sim_results} confirms our method's dominance with a 100\% SR at 300m and maintaining a highly stable 95.56\% SR at the 800m in Task 1. Furthermore, our high SPL indicates
that the generated global paths are highly efficient and strictly followed by the local executor. In contrast, NoMaD drops to 0\% SR at long range due to visual memory limits in repetitive scenes. OPEN performs adequately at 300m, while its performance degrades significantly at 800m, due to trajectory collisions over extended distances.
In Task 2, which requires the critical transition into “last-mile” open-vocabulary search, our system
sustains an 62.22\% SR at 800m, effectively leveraging the voxel-based semantic mapping to semantic targets. Meanwhile, OPEN experiences a sharp decline, which fails due to its reliance on lightweight visual encoders and a flat 2D BEV projection, where distant targets lose their 3D features and are smeared into the background. Conversely, our robust vision encoder combined with voxel-based probabilistic fusion preserves 3D structural integrity, enabling precise target isolation from clutter.

As shown in Fig. \ref{fig:rayfronts} (Top), we introduced visual ambiguity with the target while our voxel-based approach successfully distinguishes the true target.

\subsection{Real-world Deployment}
We deployed the system at a large-scale university campus to evaluate real-world performance across four 200m–500m scenarios. Table \ref{tab:ablation_rapper} reports the SPL, calculated against OSRM-generated optimal routes and GNSS-recorded actual trajectories. Crucially, ours is the only method to successfully complete all missions, whereas baselines failed due to specific limitations visualized in Fig. \ref{fig:realrobot}.
Lacking topological priors, NoMaD failed to generalize to the unseen environment, suffering from heading errors and collisions, shown in Fig. \ref{fig:realrobot}(a). Illustrated in Fig. \ref{fig:realrobot}(b), OPEN's costmap planner froze amid high-traffic dynamic obstacles. Meanwhile, illustrated in Fig. \ref{fig:realrobot}(c), its lack of heading calibration caused cumulative drift, projecting the target waypoints into physical obstacles. 
Furthermore, while OPEN reached the target area in Fig. \ref{fig:realrobot}(d), its flat 2D BEV projection and lightweight visual encoders discarded crucial spatial features, filtering the target out as noise. Conversely, our real-time 3D voxel mapping in \ref{fig:rayfronts} preserved spatial consistency, ensuring robust dynamic avoidance and precise target localization.
\begin{table}[t]
\centering
\caption{Ablation Study of RAPPER in Real-World Navigation}
\def\arraystretch{1.0}%
\setlength{\tabcolsep}{8pt}
\label{tab:ablation_rapper}
\begin{tabular}{ccccc}
\hline
 Method & Trial 1 & Trial 2 & Trial 3 & Trial 4 \\ 
\hline
N\textipa{\ae}VIS (w/o) & 0.00 & 0.00 & 21.64 &  0.00\\ 
N\textipa{\ae}VIS (w/) & 87.27 & 83.32& 86.25 & 84.28 \\ 
\hline
\end{tabular}%
\begin{tablenotes}
    \item w/o indicates without RAPPER module, and w/ indicates with RAPPER module.
\end{tablenotes}
\vspace{-20pt}
\end{table}

\subsection{Ablation Study}

To validate the necessity of RAPPER's hierarchical guidance, we conducted an ablation study by bypassing the OSM-based reasoning module on the same four scenarios (b-e) described in the real-world experiments. In the \textit{w/o}setting, we directly feed the final ENU coordinates to the high-level path planner without intermediate topological waypoints. We calculate the SPL metric shown in Table~\ref{tab:ablation_rapper}, the baseline fails in 3 out of 4 trials. Without RAPPER, the planner fails to find a feasible path in the initial open area within the timeout. Although it succeeded in one scenario, the path was highly inefficient. This confirms that topological reasoning is a prerequisite for scalable outdoor navigation.

\section{Limitations and Future Work}
Although the robot executes missions fully autonomously without real-time Internet connectivity, the current system still requires a local network link to the edge server to obtain the initial Behavior Tree. This constrains mission initiation to network-covered zones, even though subsequent execution can continue offline. Future work will explore deploying onboard LLMs to eliminate this start-up dependency and enable truly standalone task initialization.

\section{CONCLUSION}
In this work, we presented \textbf{G-DRAGON}, a unified framework capable of grounding natural language commands into large-scale outdoor navigation and ``last-mile" exploration on a local deployment. By leveraging a generative retrieval-based \textbf{GeoQA} module, we effectively bridged the gap between high-level semantic reasoning and actionable geometric goals. The integration of \textbf{RAPPER} for executable planning and \textbf{N\textipa{\ae}VIS} for open-semantic search ensures robustness against dynamic disturbances and ambiguity. Simulation trials demonstrate that our system outperforms existing baselines by 50\% and we successfully completed object search with trajectories of up to 500m on a real robot. 




\bibliographystyle{IEEEtran}
\bibliography{main}
\end{document}